\documentclass[sigconf]{acmart}
\usepackage[ruled,linesnumbered]{algorithm2e}
\usepackage{algorithmic}
\usepackage{graphicx}
\usepackage{textcomp}
\usepackage{todonotes}
\usepackage{xcolor}
\usepackage{xspace}
\usepackage{subfigure}
\usepackage{subfloat}
\usepackage{multirow}
\usepackage{wrapfig}
\usepackage{graphicx}
\usepackage{cleveref}
\usepackage{tikz}
\crefformat{subsection}{\S#2#1#3}
\crefformat{subsubsection}{\S#2#1#3}

\usepackage{array}
\usepackage{tabulary}
\newcolumntype{K}[1]{>{\centering\arraybackslash}p{#1}}
\newcolumntype{P}[1]{>{\centering\arraybackslash}p{#1}}
\newcolumntype{M}[1]{>{\centering\arraybackslash}m{#1}}

\newcommand{\squishlist}{
\begin{list}{$\bullet$}
  { \setlength{\itemsep}{0pt}
     \setlength{\parsep}{0pt}
     \setlength{\topsep}{0pt}
     \setlength{\partopsep}{0pt}
     \setlength{\leftmargin}{0em}
     \setlength{\labelwidth}{0em}
     \setlength{\labelsep}{0.2em} } }

\newcommand{\squishlisttwo}{
\begin{list}{$\bullet$}
  { \setlength{\itemsep}{0pt}
     \setlength{\parsep}{0pt}
    \setlength{\topsep}{0pt}
    \setlength{\partopsep}{0pt}
    \setlength{\leftmargin}{2em}
    \setlength{\labelwidth}{1.5em}
    \setlength{\labelsep}{0.5em} } }

\newcommand{\squishend}{
  \end{list}  }

\AtBeginDocument{
  \providecommand\BibTeX{{
    \normalfont B\kern-0.5em{\scshape i\kern-0.25em b}\kern-0.8em\TeX}}}
    
\def\BibTeX{{\rm B\kern-.05em{\sc i\kern-.025em b}\kern-.08em
    T\kern-.1667em\lower.7ex\hbox{E}\kern-.125emX}}

\setcopyright{none}
\renewcommand\footnotetextcopyrightpermission[1]{}
\acmYear{2024}\copyrightyear{2024}
\setcopyright{none}
\acmConference[Under Submission]{Paper Under Submission}{Feb 23--27, 2024}{Los Angeles, CA, USA}
\acmBooktitle{Under Submission}
\acmPrice{15.00}
\acmDOI{00.0000/0000000.0000000}
\acmISBN{000-0-0000-0000-0/00/00}

\newcommand{\name}{GDTM\xspace}

\title{\name: An Indoor Geospatial Tracking Dataset with Distributed Multimodal Sensors}

\author{Ho Lyun Jeong}
\email{holyun.lucas.jeong@gmail.com}
\authornote{The first two authors contributed equally to this paper.}
 \affiliation{
     \institution{University of California, Los Angeles}
      \city{Los Angeles} 
      \state{CA} 
      \country{USA}
      \postcode{90095}
 }

\author{Ziqi Wang}
\authornotemark[1]
 \email{wangzq312@g.ucla.edu}
 \affiliation{
     \institution{University of California, Los Angeles}
      \city{Los Angeles} 
      \state{CA} 
      \country{USA}
      \postcode{90095}
 }

\author{Colin Samplawski}
\email{csamplawski@cs.umass.edu}
 \affiliation{
     \institution{University of Massachusetts Amherst}
      \city{Amherst} 
      \state{MA} 
      \country{USA}
      \postcode{01003}
 }

\author{Jason Wu}
 \email{jaysunwu@g.ucla.edu}
 \affiliation{
     \institution{University of California, Los Angeles}
      \city{Los Angeles} 
      \state{CA} 
      \country{USA}
      \postcode{90095}
 }

\author{Shiwei Fang}
\email{shfang@augusta.edu}
 \affiliation{
     \institution{University of Massachusetts Amherst}
      \city{Amherst} 
      \state{MA} 
      \country{USA}
      \postcode{01003}
 }

\author{Lance M. Kaplan}
\email{lance.m.kaplan.civ@army.mil}
 \affiliation{
     \institution{US DEVCOM Army Research Laboratory}
      \country{USA}
 }

 \author{Deepak Ganesan}
\email{dganesan@cs.umass.edu}
 \affiliation{
     \institution{University of Massachusetts Amherst}
      \city{Amherst} 
      \state{MA} 
      \country{USA}
      \postcode{01003}
 }

\author{Benjamin Marlin}
\email{marlin@cs.umass.edu}
 \affiliation{
     \institution{University of Massachusetts Amherst}
      \city{Amherst} 
      \state{MA} 
      \country{USA}
      \postcode{01003}
 }
 
\author{Mani Srivastava}
\authornote{Mani Srivastava holds concurrent appointments as a Professor of ECE and CS (joint) at the University of California, Los Angeles and as an Amazon Scholar. This paper describes work performed at the University of California, Los Angeles and is not associated with Amazon.}
 \email{mbs@ucla.edu}
 \affiliation{
     \institution{University of California, Los Angeles}
     \institution{and Amazon}
      \country{USA}
      \postcode{90095}
 }
 
\renewcommand{\shortauthors}{Ho Lyun Jeong, Ziqi Wang, et al.}
\begin{document}

\begin{abstract}
Constantly locating moving objects, i.e., geospatial tracking, is essential for autonomous building infrastructure. Accurate and robust geospatial tracking often leverages multimodal sensor fusion algorithms, which require large datasets with time-aligned, synchronized data from various sensor types. However, such datasets are not readily available. Hence, we propose \name, a nine-hour dataset for multimodal object tracking with distributed multimodal sensors and reconfigurable sensor node placements. Our dataset enables the exploration of several research problems, such as optimizing architectures for processing multimodal data, and investigating models' robustness to adverse sensing conditions and sensor placement variances. A GitHub repository containing the code, sample data, and checkpoints of this work is available at \url{https://github.com/nesl/GDTM}.
\vspace{-7pt}

\end{abstract}

\maketitle

\section{Introduction}

Geospatial tracking has been an essential and exciting problem for autonomous building infrastructure. The task of geospatial tracking is to constantly detect and locate objects moving across a scene, providing critical localization and tracking services for mobile health, building management, surveillance, etc~\cite{zafari2019survey}. 
For example, smart building infrastructure can track the whereabouts of delivery robots to assist them with complex tasks such as elevator control that would otherwise be unfeasible for the robot to perform. First responders can also benefit from better situational awareness provided by advanced target recognition and localization services in unfamiliar indoor scenes. 

Multimodal sensor networks are particularly useful in geospatial tracking because of the GPS-denied nature of indoor environments, the complexity of layouts, and limited cooperation from the sensing targets (i.e., human beings or robots). Multimodal sensor fusion methods aggregate information from heterogeneous sensors to construct perceptions of a scene. Research has shown that multimodal sensor fusion helps improve the accuracy and reliability of sensing applications in the following ways~\cite{duan2022multimodal}. \emph{First}, multimodal sensor fusion makes systems more robust to adversarial sensing conditions, such as poor lighting conditions~\cite{cui2023radar} or bad weather~\cite{bijelic2020seeing}. \emph{Second}, multimodal sensor fusion methods have demonstrated increased resilience to unstable sources of data, i.e., missing sensor readings~\cite{john2022multimodal, piechocki2023multimodal}. \emph{Third}, the complementary sensing capability of multiple sensors also better captures the complicated nature of a scene, providing knowledge not available to unimodal sensing systems~\cite{ulrich2018person}. Despite these benefits, multimodal sensor fusion often relies on a data-driven approach, whose success depends on time-synchronized large datasets collected with a network of heterogeneous sensors.\looseness=-1

The autonomous driving community has embraced multimodal sensor fusion approaches, producing an abundance of algorithms~\cite{huang2022multi} and datasets~\cite{geiger2013vision, caesar2020nuscenes,mao2021one,sun2020scalability}, where cameras and LiDARs are the primary choice of sensors. Serving autonomous driving applications, these multimodal datasets are often collected in large outdoor spaces with mobile robots and vehicles. However, similar datasets for indoor infrastructures have been largely overlooked despite being an equally critical asset. Therefore, we focus on providing a nine-hour dataset collected with a network of multimodal sensors for the geospatial tracking problem in order to benefit the design of autonomous systems operating in indoor environments.\looseness=-1

The sensor set in \name covers most commonly seen indoor modalities, including stereo vision camera, LiDAR camera, mmWave radar, and microphone arrays. These sensors possess complementary sensing capabilities and sufficient redundancies. We created a sensor network consisting of three sensor nodes, each equipped with the aforementioned sensor set. We also designed software tools and scripts that synchronize the sensor network using Network Time Protocol (NTP), control the start and the end of data collection sessions, as well as compress and stream the data to storage devices. During the data collection, the sensor nodes are placed around an indoor race track with remote-controlled cars running on top. The sensor nodes' field-of-views are partially overlapped. \name's multi-hour data collection experiments cover the cases of a single tracking subject, multiple tracking subjects, and poor illumination. We leverage an OptiTrack motion capture system to provide mm-level ground truth for the location and orientation of the sensing target and the sensor nodes. Another highlight of \name is that the dataset is collected with a variety of different sensor placements. In most existing datasets~\cite{han2023mmptrack}, the sensors are placed at fixed positions in each scene. Models trained on such data overfit to a particular viewpoint and generalize poorly during deployments. We showcase such failures later in Fig~\ref{fig:multiview}. \name covers more than 20 different sets of sensor viewpoints to provide sufficient training and testing cases to enable the development of algorithms robust to sensor placement variations.

Our \name dataset enables the investigation of a number of research questions. For example, one may use \name to explore architectures for high-performance multimodal sensor fusion trackers that are robust to poor illumination conditions or missing sensor data. \name can also benefit research aiming to create easily deployable sensing models resilient to the placement locations and orientations of its sensor nodes. We also expect \name to benefit research of multi-object tracking, resource scheduling of sensing applications, complex event detection, and so on. In the next few sections of the paper, we first review some current literature in Sec.~\ref{Sec:related}, and then cover some details on the sensing nodes, the experiments, and statistics of \name in Sec.~\ref{Sec:dataset}. Next, we use two baseline experiments to explore a subset of the research questions raised by \name in Sec.~\ref{Sec:singleview} and Sec.~\ref{Sec:multiview}, in order to demonstrate the quality and the usefulness of \name. Finally, we discuss the limitations and future direction to wrap the paper in Sec.~\ref{Sec:final}.

\vspace{-8pt}
\section{Related Work}

\label{Sec:related}
\textbf{\textcolor{black}{Egocentric multimodal geospatial tracking datasets.}} A number of multimodal datasets have been proposed focusing on autonomous driving applications. RGB-D SLAM Dataset~\cite{sturm12iros} places RGB-D cameras on a robot moving in the scene and uses an OptiTrack motion capture system to provide ground truth. \textcolor{black}{Indoor Location Competition 2.0 Dataset~\cite{shu2021indoor} includes IMU, Wi-Fi, and Bluetooth data collected from a user-held cellphone and uses pre-defined waypoints as the ground truth for indoor location sensing tasks.} 
StreetAware~\cite{piadyk2023streetaware} combines video, audio, and LiDAR to construct a multimodal urban scene dataset. Other datasets like KITTI~\cite{geiger2013vision}, nuScenes~\cite{caesar2020nuscenes}, ONCE~\cite{mao2021one}, and Waymo Open~\cite{sun2020scalability} also fall into this category, focusing on large outdoor spaces for developing self-driving algorithms. \textcolor{black}{These datasets typically operate with mobile vehicles and sense the environment in an ego-centric manner. Our \name dataset, on the other hand, employs a network of indoor multimodal sensors to sense non-cooperative targets to investigate the research opportunities provided by smart building infrastructures.}\looseness=-1

\textbf{Datasets collected with sensor networks.} Meanwhile, there exists a number of datasets that focus on sensor networks where perception happens across different sensors and viewpoints. The VIRAT video dataset~\cite{oh2011large} employs multiple video surveillance cameras to provide benchmarks for human activity recognition. A follow-up dataset work, MEVA~\cite{Corona_2021_WACV}, extends the dataset to more camera views, including some infrared cameras and drone-mounted cameras, and uses GPS to provide the ground truth of different person's location. MMPTRACK~\cite{han2023mmptrack} aims to create a large-scale tracking dataset using a multi-camera tracking system for multiple human subjects. Radio-frequency wireless signals are also commonly used in indoor environments since they are less susceptible to light conditions and occlusions. OperaNet~\cite{bocus2022operanet} combines Kinect visual sensors with multiple radio-frequency wireless sensing modalities (e.g., Wi-Fi channel state information) to construct a multimodal activity recognition dataset. 
\textcolor{black}{Compared with existing dataset work}, \name differs both in the richness of the sensors and the diversity of viewpoints. Sensor nodes of \name cover vision, depth, radar, and audio, which are the most commonly seen indoor sensing modalities. 
Moreover, \name provides multiple sets of sensor placements, encouraging the development of fusion models robust to the domain shift caused by changed sensor perspectives.\looseness=-1

\section{Dataset Collection}

\label{Sec:dataset}
\subsection{Sensing Platform}

\begin{figure}[t!]
    \centering
        \vspace{-10pt}
    \includegraphics[width=1.00\linewidth]{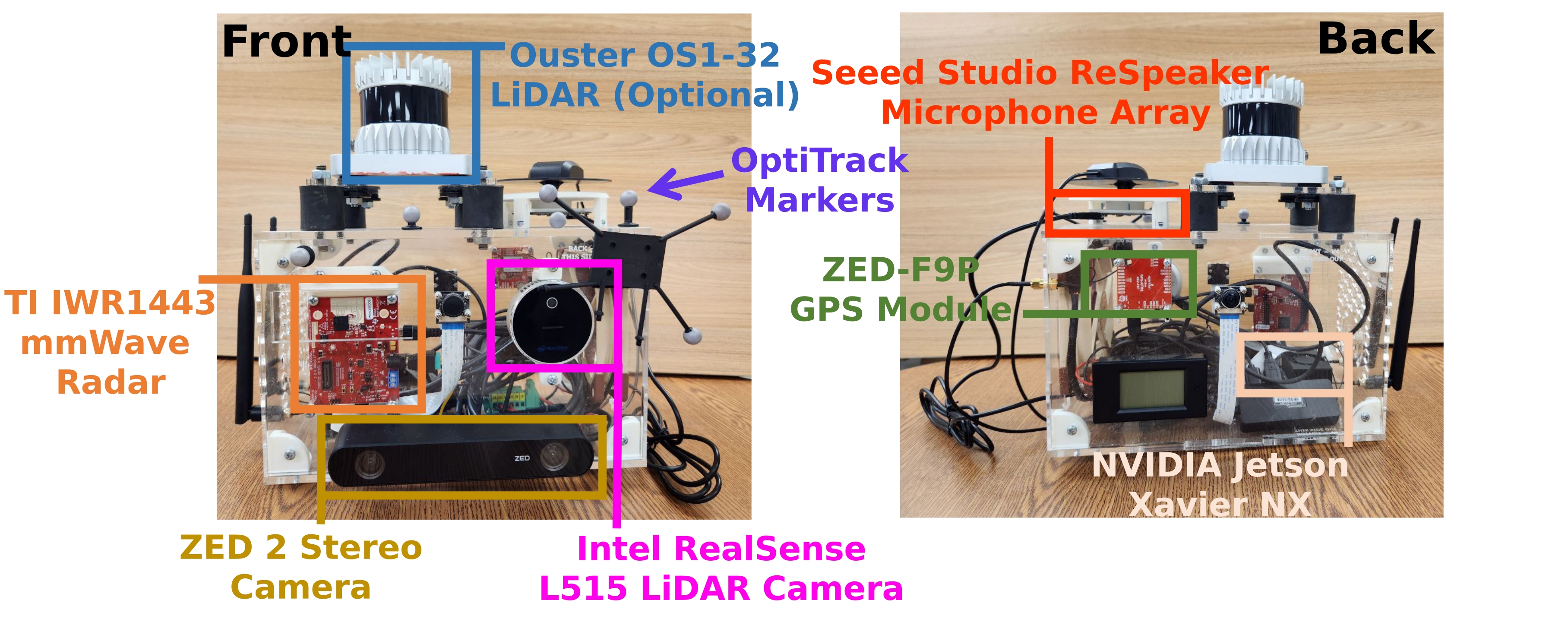}
    \vspace{-20pt}
    \caption{One of our multimodal sensor data collection nodes.}
    \label{fig:node}.
        \vspace{-10pt}
\end{figure}

We designed and assembled a multimodal sensing platform consisting of three sensor nodes. We showcase one of the sensor nodes in Figure~\ref{fig:node}. Each node is equipped with a complementary sensor set, including a RealSense L515 LiDAR camera, a ZED 2i stereo vision camera, an IWR1443 mmWave Radar, and a ReSpeaker microphone array. To cater to the possibility of outdoor operation, one of the sensing nodes contains a commercial LiDAR, and all three nodes have a GPS-RTK board. An NVIDIA Jetson Xavier NX serves as the main controller of each sensing node, responsible for controlling sensor operations, maintaining node health information, temporarily storing the data, and listening to the instructions from a command node.

\begin{figure}[t!]
    \centering
    \includegraphics[width=1.00\linewidth]{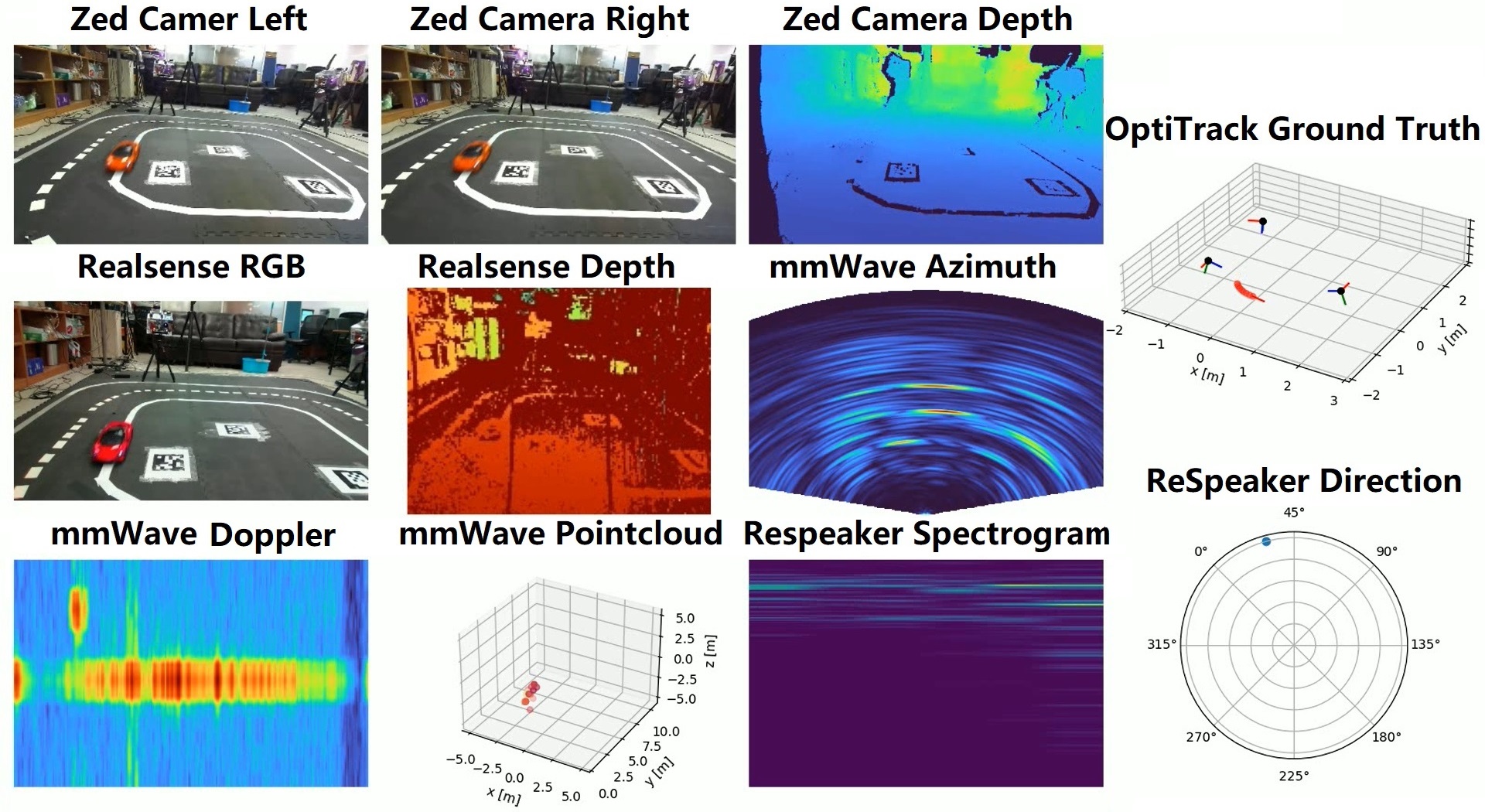}
    \vspace{-20pt}
    \caption{Sample data collected by one of the 3 sensing nodes.}
    \vspace{-15pt}
    \label{fig:data}
\end{figure}

\begin{figure*}[t!]
    \centering
    \includegraphics[width=0.65\linewidth]{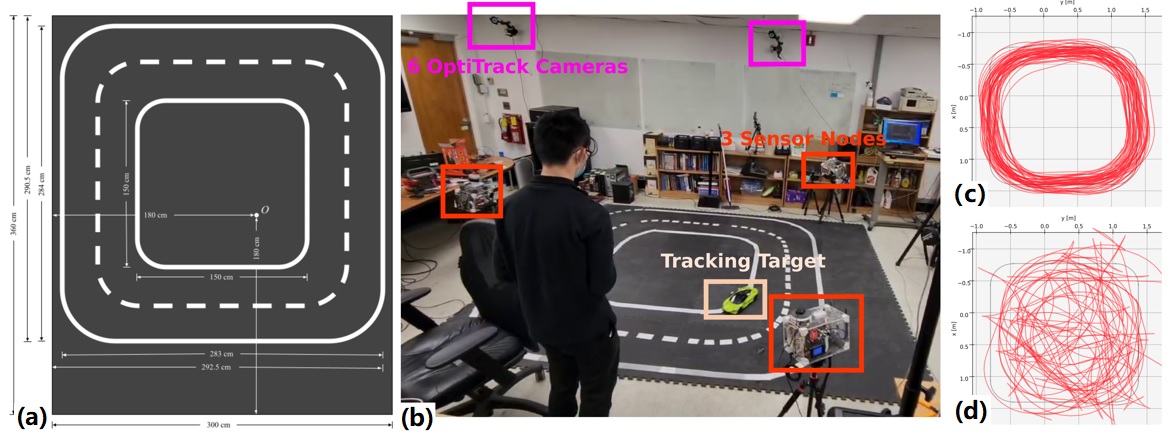}
    \vspace{-10pt}
    \caption{\textcolor{black}{Data collection session setup: (a) racetrack dimensions (b) experiment site (c) an exemplar circular trajectory (d) an exemplar random trajectory.}}
    \vspace{-15pt}
    \label{fig:setting}
\end{figure*}
A small NUC Mini computer serves as the command node. The command node and three sensor nodes are placed under the same Wi-Fi network and time-synchronized using the NTP protocol. During the experiment, the command node sends instructions to the sensor nodes to control the start and the end of sensor data collection. The command node also acquires each sensing node's health status during the experiments to ensure data continuity. At the end of each experiment session, the acquired data will be compressed and streamed to the command node. 

Sample data from one of the sensor nodes is shown in Figure~\ref{fig:data}. In \name, we provide both left and right images, as well as the calculated depth from the stereo vision camera at 15 Hz with $1920 \times 1080$ resolution. From LiDAR cameras, we obtain RGB frames captured at 15 Hz with $1920 \times 1080$ and corresponding depth frames streamed at 15 Hz with $640 \times 480$ resolution. 
From the mmWave radars, range azimuth heatmap, range doppler heatmap, range profile, noise profile, and point cloud are recorded at 4 Hz. The six audio channels of ReSpeaker (raw data from the four microphones, merged playback, and processed audio for automatic speech recognition, sampled at 16~\!kHz) are encoded in FLAC format. While not being a main focus of this paper, Ouster LiDAR scans are also included in \name. The ground truth contains mm-level pose information (position and orientation) for each node and each car on the track collected by the OptiTrack motion capture system, which tracks the reflective balls mounted on the tracked subjects and the sensor nodes using infrared light.\looseness=-1

\vspace{-15pt}
\subsection{Experiment Setup}

\noindent\textbf{Experiment Site.} The experiments are conducted in an indoor race track that spans an area of $3 \textrm{ m} \times 3.6 \textrm{ m}$. The race track is delineated with white tape on top of dark gray foam mats. The detailed measurements of the race track are depicted in Figure ~\ref{fig:setting}(a). Six OptiTrack cameras are set up around the room to collect ground truth. Sensor nodes are placed around the race track during each experiment. Figure~\ref{fig:setting}(b) shows the experiment site.

\noindent\textbf{Experiment Settings.} We pick remote-controlled cars (RC cars) as targets for geospatial tracking in \name since RC cars can easily create arbitrary trajectories we desire. For each experiment, we drive a red RC car, a green RC car, or both around the race track. The trajectory of each car mostly follows the lanes on the race track, with behaviors like changing lanes, turning around, or backing up. To avoid creating monotonous circular trajectories, we also drove RC cars in random or erratic patterns. This random behavior allows for verification that the tracking system is not simply predicting repetitive motion. \textcolor{black}{Figure~\ref{fig:setting}(c) shows a circular trajectory, and Figure~\ref{fig:setting}(d) demonstrates a random trajectory. Each trajectory is plotted for a duration of 200 seconds. We quantify the motion patterns by dividing the trajectories into chunks of 10 seconds and classifying them based on whether the chunk dominantly lies within the racetrack lanes. We found that 80.8\% of our trajectories are circular patterns, and the remaining 19.2\% are random patterns. 
Note that even in the circular patterns, the trajectories are not simply repeating themselves but are actually ``donut-shaped'', covering a width of 60-80 cm, allowing for greater diversity.
Thus, combining both random and circular patterns, we have good coverage of the testing site.}\looseness=-1

\name also covers the case of multi-object tracking by including over one hour of data with both cars on the race track, featuring complex dynamics such as overtakes and collisions. Meanwhile, \name contains around three hours of data under poor lighting conditions. This setting challenges the robustness of the tracking system as the utility of certain modalities, such as RGB and stereo vision, is vastly reduced. Furthermore, \name features numerous positional configurations (positions and orientations) of the three collection nodes, including over 15 different views for experiments with each car in good lighting. This prevents the tracking system from overfitting to a fixed viewpoint, encouraging it to learn a representation more generalizable to different deployments. A detailed breakdown of the dataset statistics can be found in Table~\ref{tab:stats}.

\begin{table}[bthp!]
\centering

\vspace{-10pt}
 \resizebox{\linewidth}{!}{  
\begin{tabular}{|c|ccc|cc|}
\hline
\multirow{2}{*}{}  & \multicolumn{3}{c|}{Good Lighting Condition}                         & \multicolumn{2}{c|}{Poor Lighting Condition} \\ \cline{2-6} 
                   & \multicolumn{1}{c|}{Red Car} & \multicolumn{1}{c|}{Green Car} & Both & \multicolumn{1}{c|}{Red Car}   & Green Car   \\ \hline
\# of Viewpoints   & \multicolumn{1}{c|}{18}      & \multicolumn{1}{c|}{15}        & 6    & \multicolumn{1}{c|}{7}         & 2           \\ \hline
Minutes            & \multicolumn{1}{c|}{200}     & \multicolumn{1}{c|}{165}       & 95   & \multicolumn{1}{c|}{145}       & 35          \\ \hline
Raw Data Size (GB) & \multicolumn{1}{c|}{278}     & \multicolumn{1}{c|}{235}       & 134  & \multicolumn{1}{c|}{202}       & 45          \\ \hline
\end{tabular}
}
\caption{Dataset Statistics. \label{tab:stats}}
\end{table}

\vspace{-30pt}
\section{Benchmarking Experiments}

\begin{table*}[t!]

\resizebox{0.7\linewidth}{!}{
\begin{tabular}{|c|cccccc|cccccc|}
\hline
\multirow{3}{*}{\begin{tabular}[c]{@{}c@{}}Fusion\\ Model\\ (Sensors)\end{tabular}} & \multicolumn{6}{c|}{Good Light Condition}                                                                                                                                                               & \multicolumn{6}{c|}{Poor Light Condition}                                                                                                                                                                \\ \cline{2-13} 
                                                                                    & \multicolumn{2}{c|}{View 1}                                              & \multicolumn{2}{c|}{View 2}                                             & \multicolumn{2}{c|}{View 3}                        & \multicolumn{2}{c|}{View 1}                                             & \multicolumn{2}{c|}{View 2}                                              & \multicolumn{2}{c|}{View 3}                         \\ \cline{2-13} 
                                                                                    & \multicolumn{1}{c|}{AD}             & \multicolumn{1}{c|}{NLL}           & \multicolumn{1}{c|}{AD}            & \multicolumn{1}{c|}{NLL}           & \multicolumn{1}{c|}{AD}            & NLL           & \multicolumn{1}{c|}{AD}            & \multicolumn{1}{c|}{NLL}           & \multicolumn{1}{c|}{AD}             & \multicolumn{1}{c|}{NLL}           & \multicolumn{1}{c|}{AD}             & NLL           \\ \hline
\begin{tabular}[c]{@{}c@{}}Early\\ (Camera)\end{tabular}                            & \multicolumn{1}{c|}{25.85}          & \multicolumn{1}{c|}{7.81}          & \multicolumn{1}{c|}{9.11}          & \multicolumn{1}{c|}{5.79}          & \multicolumn{1}{c|}{14.69}         & 7.03          & \multicolumn{1}{c|}{117.14}        & \multicolumn{1}{c|}{15.49}         & \multicolumn{1}{c|}{118.91}         & \multicolumn{1}{c|}{16.4}          & \multicolumn{1}{c|}{115.94}         & 15            \\ \hline
\begin{tabular}[c]{@{}c@{}}Early\\ (All)\end{tabular}                               & \multicolumn{1}{c|}{\textbf{16.26}} & \multicolumn{1}{c|}{\textbf{7.29}} & \multicolumn{1}{c|}{8.85}          & \multicolumn{1}{c|}{6.08}          & \multicolumn{1}{c|}{\textbf{6.25}} & \textbf{5.26} & \multicolumn{1}{c|}{\textbf{19.9}} & \multicolumn{1}{c|}{\textbf{8.14}} & \multicolumn{1}{c|}{19.89}          & \multicolumn{1}{c|}{\textbf{8.13}} & \multicolumn{1}{c|}{\textbf{16.21}} & \textbf{7.26} \\ \hline
\begin{tabular}[c]{@{}c@{}}Late\\ (Camera)\end{tabular}                             & \multicolumn{1}{c|}{23.79}          & \multicolumn{1}{c|}{7.63}          & \multicolumn{1}{c|}{\textbf{5.92}} & \multicolumn{1}{c|}{\textbf{5.36}} & \multicolumn{1}{c|}{10.16}         & 6.25          & \multicolumn{1}{c|}{115.2}         & \multicolumn{1}{c|}{163.27}        & \multicolumn{1}{c|}{45.94}          & \multicolumn{1}{c|}{10.83}         & \multicolumn{1}{c|}{32.73}          & 8.3           \\ \hline
\begin{tabular}[c]{@{}c@{}}Late\\ (All)\end{tabular}                                & \multicolumn{1}{c|}{31.93}          & \multicolumn{1}{c|}{19.86}         & \multicolumn{1}{c|}{12.14}         & \multicolumn{1}{c|}{9.7}           & \multicolumn{1}{c|}{9.22}          & 6.25          & \multicolumn{1}{c|}{47.18}         & \multicolumn{1}{c|}{34.95}         & \multicolumn{1}{c|}{\textbf{16.08}} & \multicolumn{1}{c|}{8.33}          & \multicolumn{1}{c|}{24.97}               & 10.28     \\ \hline
\end{tabular}}
\caption{Comparison of four variants of multimodal sensor fusion architectures in both good and poor illuminations.} \label{tab:exp1}
\vspace{-20pt}
\end{table*}

We designed two benchmarking experiments to showcase two of many research questions arising from the \name dataset. \textcolor{black}{Specifically, we focused on improving geospatial tracking accuracy by investigating model architectures for sensor fusion, and increasing tracking robustness by creating models robust to the domain shift caused by sensor perspective changes.}\looseness=-1

\vspace{-5pt}
\subsection{Baseline 1: Exploring Multimodal Sensor Fusion Architectures}
\label{Sec:singleview}
\vspace{-3pt}
In this section, we compare an early fusion and a late fusion approach by evaluating their tracking performance in different lighting conditions.\looseness=-1

\vspace{-3pt}
\subsubsection{Model Architectures:}
\vspace{-3pt}
\begin{figure}[t!]
    \centering
    \includegraphics[width=1.00\linewidth]{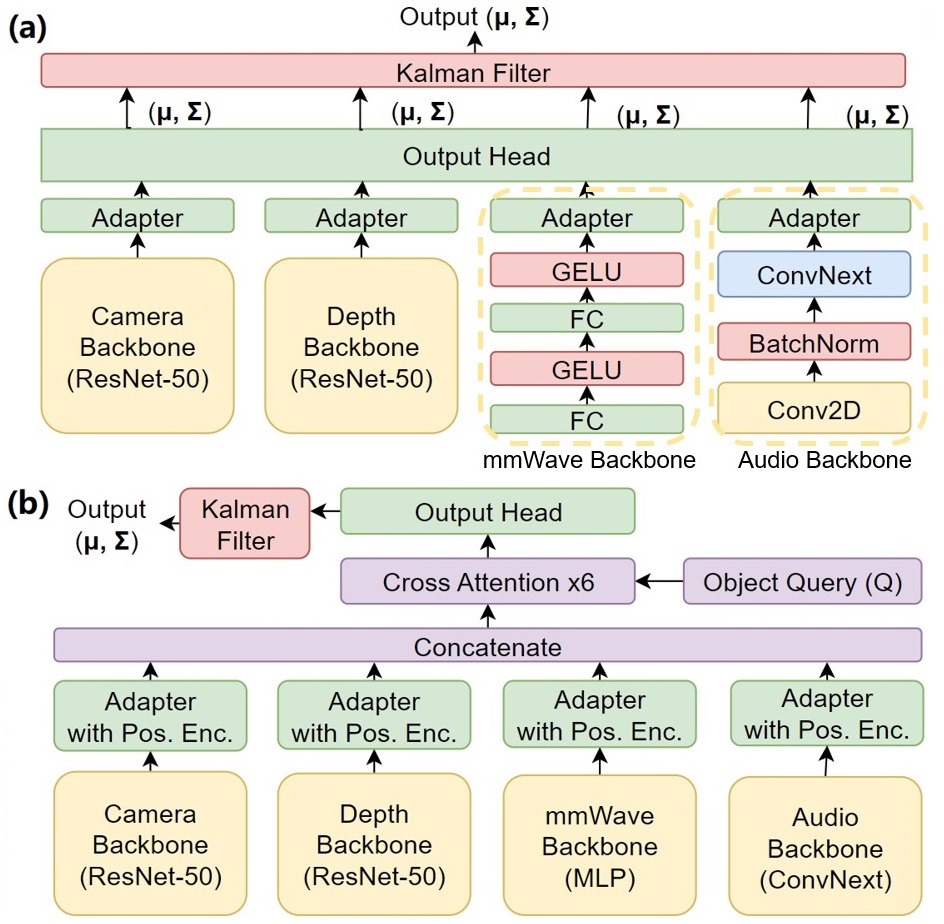}
    \vspace{-10pt}
    \caption{\textcolor{black}{Multimodal sensor fusion architecture.}}
    \vspace{-20pt}
    \label{fig:architecture}
\end{figure}
\noindent\textbf{Late Fusion Model Architecture.} Video-based object tracking has been a well-investigated area where mature architectures such as mmtracking~\cite{mmtrack2020} have been proposed. In this first experiment, we aim to extend the video-based tracking architectures to multimodal sensor data. Specifically, we adopt the heteroskedastic geospatial detector (HGD) structure proposed in~\cite{samplawski2020heteroskedastic}, as shown in Figure~\ref{fig:architecture}(a). HGD consists of a backbone that extracts features from raw sensor data, an adapter (a fully-connected layer) that maps the extracted features to a fix-sized ($1\times256$) shared embedding, and an output head that converts the latent encoding into the object's mean location and covariance using another fully-connected layer. HGD then uses a Kalman filter to reconcile the predictions from different branches and output a final distribution of the objects' predicted locations. In HGD, the backbone is shared within each modality, and the output head is shared by all nodes and all modalities. Adapters, however, are unique to the combination of nodes and modalities. With 3 nodes and 4 modalities, we have four backbones, 12 adapters, and one output head.\looseness=-1 

We extend the HGD architecture by designing backbones for all modalities. For vision-based modalities such as RGB and depth, we use ResNet-50 pre-trained on ImageNet as the backbone. For radar, we use an MLP with two fully connected layers interleaved with GELU non-linearity layers. For audio, we stack a 2D convolution, a BatchNorm layer, and finally, a ConvNext layer~\cite{liu2022convnet} as the backbone. The benefit of using HGD is that it accounts for prediction uncertainties by outputting a multivariate normal distribution rather than a single location prediction.

\textbf{Early Fusion Model Architecture.} In the late fusion model above, each modality generates its own predictions that are later reconciled by a Kalman filter. To facilitate early information exchange across sensors, we propose an early fusion approach where the fusion happens at the intermediate feature level (see Figure~\ref{fig:architecture}(b))
The inputs are first processed by the same backbones used in the late fusion model. 

Then, the adapter (with positional encoding) flattens the backbone features into $D_k \times 256$ feature vectors, where $D_k$ is related to the original size of the $k_{th}$ input branch. Later in the model, we will use an attention mechanism that ignores the order of the features, which results in the loss of spatial information. For example, the camera backbone extracts features from the input image, where each feature represents a pixel region in the original image. These features are ordered sequentially to preserve the spatial information. The attention mechanism, however, is agnostic to feature ordering and cannot retain spatial information. To address this issue, we apply sigmoid positional encodings proposed by~\cite{carion2020end}. We calculate an embedding vector $\mathbf{E}$ of size 256 representing the spatial order of each feature vector $\mathbf{F}$, and then append the embedding $\mathbf{E}$ to $\mathbf{F}$, increasing the feature dimension from 256 to 512. The appending operation is inspired by~\cite{meng2021conditional} to reduce the likelihood that the model will learn to ignore the positional encodings.

The outputs of these adapters are concatenated into a latent feature vector $\mathbf{V} \in \mathrm{R}^{D \times 512}$, where $D = \sum_k d_k$. This is where the \textbf{fusion} begins. The latent feature vector is then fed to a six-layer cross-attention module. Each layer is an 8-head cross-attention accepting three sets of vectors (query, key, and value) as inputs.
Similar to \cite{carion2020end}, we instantiate a single $512$-dim query embedding $\mathbf{Q}$ for the first cross-attention layer. This embedding is a set of learnable parameters that encode the latent prediction of the object's location. We update the embedding by applying repeated layers of cross-attention. Here, the query and key of the attention come from $\mathbf{Q}$, and the value is calculated from the latent feature vector $\mathbf{V}$ introduced above. 
The output of each cross-attention layer is still $1\times512$, which becomes the $\mathbf{Q}$ for the next attention layer. 
The final embedding is then decoded into a Gaussian distribution using the output head of the model, followed by a Kalman filter similar to the one used in the late fusion architecture. To support multi-object tracking in the future, we can instantiate more $\mathbf{Q}$ vectors, with each of them being responsible for one object.\looseness=-1



In both models, the loss is defined by the negative log-likelihood (NLL) of the ground truth position of the object under the distribution specified by the output head. After training, some additional data is required to perform a post-hoc model recalibration as described in \cite{samplawski2020heteroskedastic} to better capture model prediction uncertainties. More specifically, we apply an affine transformation $\Sigma' = a \Sigma + b I$ to the output covariance matrix $\Sigma$ with parameters $a$ and $b$ that minimizes the calibration data's NLL. \textcolor{black}{Our models process the sensor data and make predictions frame-by-frame. The new predictions are reconciled with each other and with the historical predictions using a Kalman tracker. }\looseness=-1

\vspace{-5pt}
\subsubsection{Evaluation metrics:}
\vspace{-3pt}

\textbf{Average Distance (AD).} We use the average distance (AD) metric to quantify the discrepancy between the model prediction and the ground truth trajectory of the object. The average distance is computed by taking the mean of the distance between the predicted and the ground truth location of the object at each frame. A lower AD value indicates a higher accuracy and a closer alignment between the predicted and the actual trajectories.

\textbf{Negative Log-Likelihood (NLL).} We use the negative log-likelihood (NLL) metric to quantify the ability of the models to probabilistically predict the trajectory of the object with a multivariate Gaussian distribution. A lower NLL value indicates a higher similarity between the model's predicted distribution and the actual ground truth distribution.

\vspace{-5pt}
\subsubsection{Experiments and Results}\label{sec:base1-exp}
\vspace{-3pt}

Both the early and late fusion models are trained, calibrated, and tested on a total of six different sets of data selected from \name, three sets under good lighting conditions and three under poor lighting conditions. Each set consists of approximately 10 minutes (9000 frames) for training, 1 minute (900 frames) for calibration, and 3 minutes (2700 frames) for testing. The training of each set takes about six hours on a single RTX A6000 GPU. Note that, in this section, the training and testing are conducted using data from different sessions but from the same viewpoints (sensor placements).\looseness=-1

\begin{figure}[b!]
    \centering
    \vspace{-12pt}
    \includegraphics[width=\linewidth]{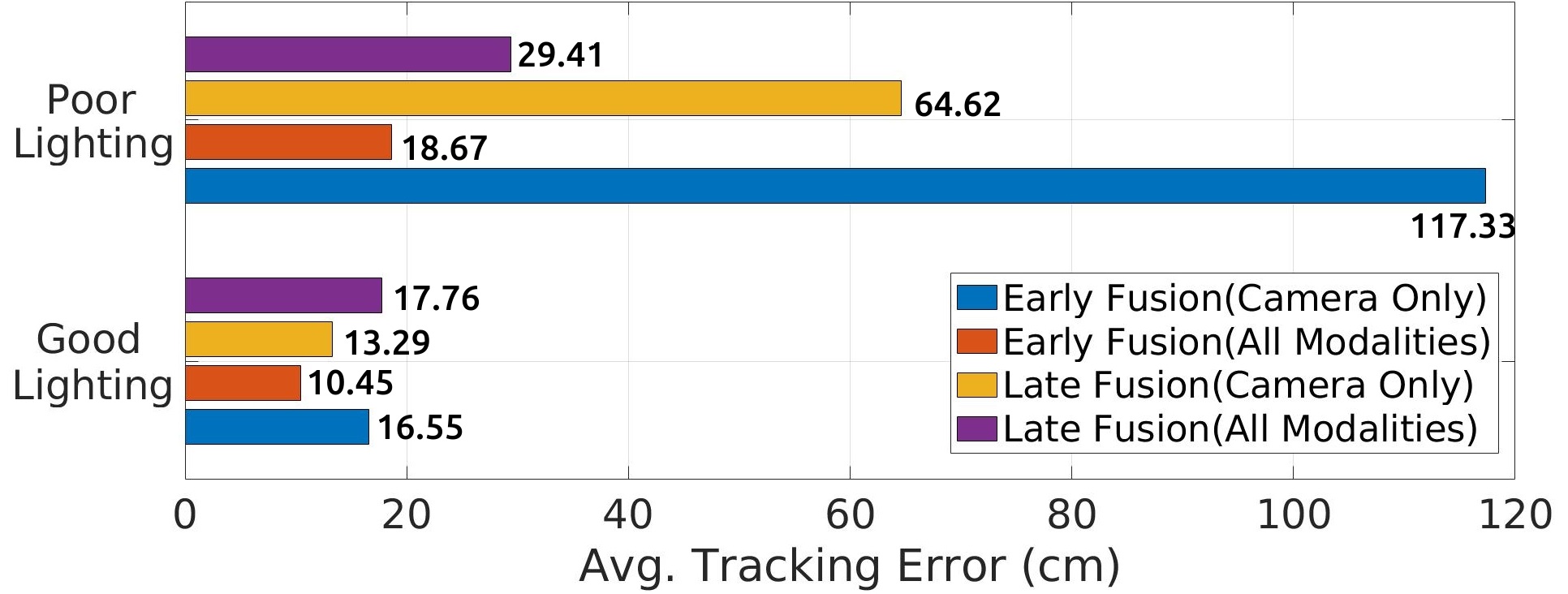}
    \vspace{-13pt}
    \caption{Average tracking error of the four model variants under different lighting conditions.}
    \label{fig:ad}
\end{figure}

\begin{figure}[b!]
    \centering
    \includegraphics[width=\linewidth]{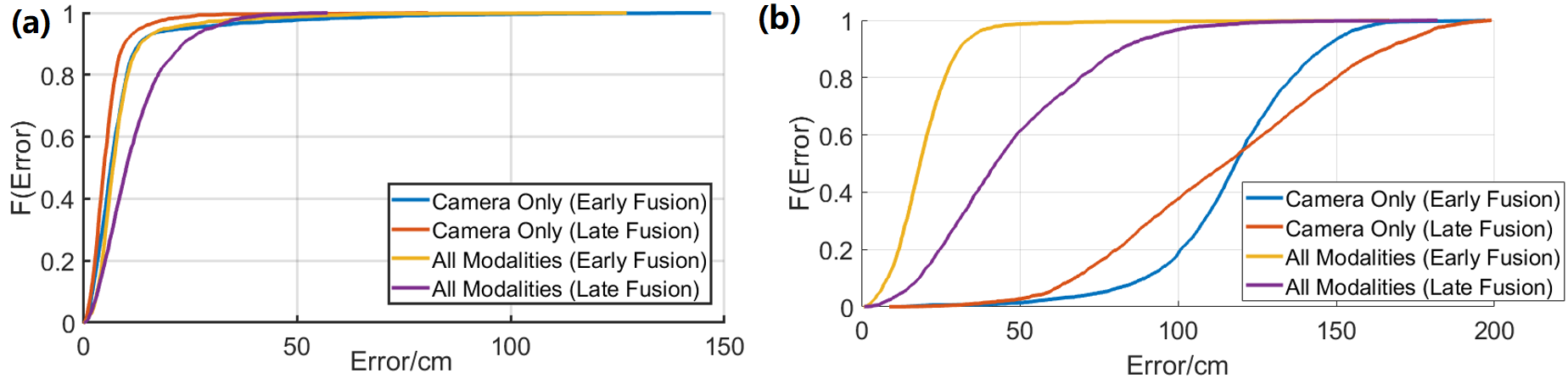}
    \vspace{-25pt}
    \caption{\textcolor{black}{Cumulative distribution function (CDF) plot of different models' tracking error.}}
    \label{fig:cdf}
\end{figure}

We conduct tests on the two models, each with two configurations: RGB camera only and all modalities combined. We investigate the effect of environmental challenges on single- and multi-modality models by subjecting them to poor lighting conditions  (see results in Table~\ref{tab:exp1}). Generally, the all-modality version of the early fusion model outperforms other variants. We then average the results across experiments to generate Figure~\ref{fig:ad}. We can see that in good lighting conditions, both the camera-only and all-modality models perform well since the camera feed contains sufficient information about the car's location. However, our camera-only models struggle under poor lighting conditions, with average tracking errors (AD) exceeding 100~\!cm for early fusion and AD=64.62~\!cm for late fusion. When additional modalities are incorporated, the performance improves significantly, reducing the AD to approximately 18 cm (using early fusion).  
\textcolor{black}{Additionally, we show the cumulative distribution function (CDF) plots of different models' positioning errors in Figure~\ref{fig:cdf}. The CDF plots clearly show that multimodal fusion methods outperform their camera-only counterparts, especially early fusion architecture in poor illumination conditions.}
These findings emphasize the role of multimodality in improving model robustness under environmental constraints, and highlight the usefulness of the \name dataset for evaluating multimodal models.\looseness=-1

\begin{table}[t!]
\centering
\resizebox{\linewidth}{!}{
\begin{tabular}{|c|cc|cc|}
\hline
\multirow{2}{*}{\begin{tabular}[c]{@{}c@{}}Enabled\\ Embeddings\end{tabular}} & \multicolumn{2}{c|}{Good Light Condition} & \multicolumn{2}{c|}{Poor Light Condition} \\ \cline{2-5} 
                                                                              & \multicolumn{1}{c|}{AD↓}       & NLL↓     & \multicolumn{1}{c|}{AD↓}       & NLL↓     \\ \hline
Camera Only                                                                   & \multicolumn{1}{c|}{16.15}     & 7.79     & \multicolumn{1}{c|}{116.27}    & 37.04    \\ \hline
Depth Only                                                                    & \multicolumn{1}{c|}{118.21}    & 40.69    & \multicolumn{1}{c|}{117.06}    & 37.48    \\ \hline
Doppler Only                                                                  & \multicolumn{1}{c|}{25.85}     & 22.84    & \multicolumn{1}{c|}{17.06}     & 7.03     \\ \hline
Audio Only                                                                    & \multicolumn{1}{c|}{89.8}      & 30.07    & \multicolumn{1}{c|}{100.76}    & 25.19    \\ \hline
All Modalities                                                                & \multicolumn{1}{c|}{6.25}      & 5.26     & \multicolumn{1}{c|}{16.21}     & 7.26     \\ \hline
\end{tabular}
}
\caption{Early fusion model performance when the embeddings of all-but-one modality are nullified.}
\label{tab:contrib}
\vspace{-20pt}
\end{table}

Furthermore, we investigate each modality's contribution toward the final prediction in the early fusion model by nullifying other modalities' embedding. In Figure~\ref{fig:architecture}(b), the latent embedding is constructed by concatenating the adapter outputs of each modality. We first train and freeze the early fusion model, set the non-target modalities' embedding to zero, and then evaluate the model performance. In this case, only the target modality contributes to the final predictions. The result (see Table~\ref{tab:contrib}) shows that the early fusion model ``pays more attention'' to the camera and Doppler information under good illuminations and focuses on mmWave Doppler information when lighting is poor. It also reveals that early fusion models with attention tend to rely on a small subset of modalities that provide sufficient information, necessitating techniques such as regularization to ensure model redundancy.\looseness=-1

\vspace{-3pt}
\subsection{Baseline 2: Developing Models Robust to Sensor Placements}
\label{Sec:multiview}

\begin{table*}[t!]

\resizebox{0.80\linewidth}{!}{
\begin{tabular}{|c|cc|cc|cc|cc|cc|cc|cc|}
\hline
                                                                              & \multicolumn{2}{c|}{View1}                                                       & \multicolumn{2}{c|}{View2}                                                                         & \multicolumn{2}{c|}{View3}                                                                         & \multicolumn{2}{c|}{View4}                                                                         & \multicolumn{2}{c|}{View5}                                                                & \multicolumn{2}{c|}{View6}                                                                & \multicolumn{2}{c|}{{ Average}}                                                \\ \hline
\begin{tabular}[c]{@{}c@{}}Model\\ (Sensors)\end{tabular}                     & \multicolumn{1}{c|}{AD}                           & NLL                          & \multicolumn{1}{c|}{AD}                                    & NLL                                   & \multicolumn{1}{c|}{AD}                                    & NLL                                   & \multicolumn{1}{c|}{AD}                                    & NLL                                   & \multicolumn{1}{c|}{AD}                                    & NLL                          & \multicolumn{1}{c|}{AD}                                    & NLL                          & \multicolumn{1}{c|}{{ AD}}             & { NLL}            \\ \hline
\begin{tabular}[c]{@{}c@{}}Late3D\\ (Camera)\end{tabular}                     & \multicolumn{1}{c|}{124.06}                       & 94.73                        & \multicolumn{1}{c|}{114.98}                                & 79.6                                  & \multicolumn{1}{c|}{136.16}                                & 115.65                                & \multicolumn{1}{c|}{133.28}                                & 78                                    & \multicolumn{1}{c|}{117.34}                                & 61.96                        & \multicolumn{1}{c|}{119.95}                                & 87.62                        & \multicolumn{1}{c|}{{ 124.30}}         & { 86.26}          \\ \hline
\begin{tabular}[c]{@{}c@{}}Late3D\\ (Camera,\\ Depth)\end{tabular}            & \multicolumn{1}{c|}{\textbf{42.53}}               & \textbf{15.72}               & \multicolumn{1}{c|}{54.64}                                 & 17.92                                 & \multicolumn{1}{c|}{85.15}                                 & 50.84                                 & \multicolumn{1}{c|}{53.72}                                 & 31.89                                 & \multicolumn{1}{c|}{74.18}                                 & \textbf{48.81}               & \multicolumn{1}{c|}{54.83}                                 & \textbf{34.85}               & \multicolumn{1}{c|}{{ 60.84}}          & { \textbf{33.34}} \\ \hline
{ \begin{tabular}[c]{@{}c@{}}Late3D\\ (All)\end{tabular}} & \multicolumn{1}{c|}{{ 45.00}} & { 31.32} & \multicolumn{1}{c|}{{ \textbf{33.71}}} & { \textbf{17.64}} & \multicolumn{1}{c|}{{ \textbf{61.89}}} & { \textbf{41.11}} & \multicolumn{1}{c|}{{ \textbf{34.07}}} & { \textbf{19.27}} & \multicolumn{1}{c|}{{ \textbf{67.45}}} & { 82.97} & \multicolumn{1}{c|}{{ \textbf{42.77}}} & { 36.05} & \multicolumn{1}{c|}{{ \textbf{47.48}}} & { 38.06}          \\ \hline
\end{tabular}}
\caption{Performances of models trained and tested on a variety of sensor placements.}
\label{tab:exp2}
\vspace{-20pt}
\end{table*}

\textcolor{black}{In this section, we showcase \name's potential to be used for solving the domain adaptation problem caused by changed sensor perspectives during deployment. Domain adaptation refers to leveraging labeled data in one or more related source domains to learn a model for unseen or unlabeled data in a target domain~\cite{csurka2017domain}. In our application scenarios, a neural network can be developed and trained with a number of camera/sensor perspectives. However, during deployment, the cameras/sensors’ locations and poses are different from that of the training data, and the network needs to be tuned or adapted before it can work properly. In this case, the source and target domain consist of sensor data coming from different perspectives.}

\begin{figure}[h!]
\vspace{-10pt}
    \centering
    \includegraphics[width=0.98\linewidth]{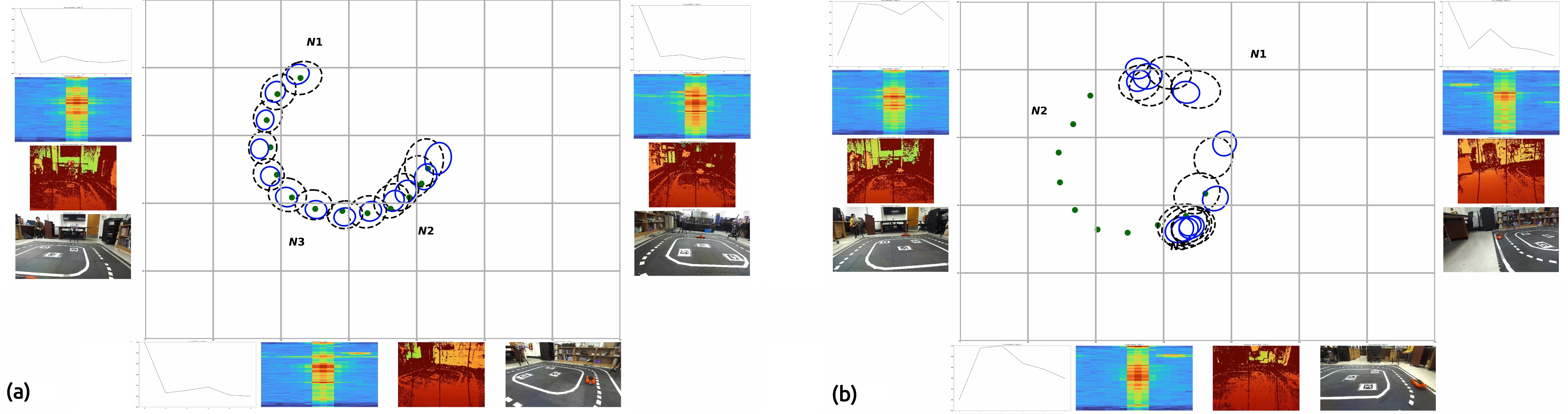}
    \vspace{-10pt}
    \caption{Model performance on (a) trained viewpoint (b) unseen viewpoint.}
    \vspace{-10pt}
    \label{fig:multiview}

\end{figure}

\textcolor{black}{We use the models developed in Section~\ref{Sec:singleview} to show the domain shift problem intuitively. In Figure~\ref{fig:multiview}(a), the green dots stand for ground-truth target locations, black circles represent the early fusion model's predictions, and the blue ovals represent the final Kalman tracker's predictions. The center of the ovals is the predicted target location, and the size of the ovals stands for prediction covariance (uncertainties). We find that the model accurately tracks the targets with data from the viewpoint on which the model was trained (i.e., source domain data). However, in Figure~\ref{fig:multiview}(b), we find that the model fails catastrophically when trying to track the object using target domain data (i.e., data from a sensor viewpoint that the model has never seen before).}

In Section~\ref{Sec:singleview}, the models were trained on a set of \emph{global} coordinates, where the models learn the correspondences between sensor readings and objects' locations. However, these correspondences break up when the sensors move to a new perspective. \textcolor{black}{To provide a preliminary baseline addressing this challenge}, we instead train nodes to localize the object within \emph{its own local coordinate system} and transform the predictions to global coordinates using sensor node pose information.\looseness=-1

\noindent\textbf{Late Fusion Model Modifications:} The late fusion model is more suitable for this task as we need to make predictions for each node and transform the result separately. A few modifications need to be made to the model for this new task. Firstly, we need to enable 3D predictions for the model. In Sec.~\ref{Sec:singleview}, the cars move on the ground plane ($xz$-plane, $y = 0$), so the coordinates are two-dimensional. These coordinates become 3D when transformed into the nodes' local coordinates \textcolor{black}{because sensor nodes are not placed parallel to the ground plane}. Thus, we must modify the \emph{Output Head} in Fig.~\ref{fig:architecture} by increasing the output dimension from 5 (2 for mean and 3 for covariance matrix) to 9 (3 for mean and 6 for covariance matrix).

We also modify the \emph{adapter} from being independent for each node to being \emph{shared} across nodes. These adapters previously generated embeddings that allowed each node to predict the targets' global coordinates, so they were independent as each node learned a different mapping. However, each node is now predicting in its own local coordinate, so the adapter can be shared \textcolor{black}{because all the nodes are built from the same blueprint and share the same layout.}\looseness=-1

\noindent\textbf{Coordinate Transformations:} \name utilizes the OptiTrack motion capture system to record the object's global coordinates while also recording the transformation matrix between each node's local coordinate system to the global. To compute the loss for training, we must provide ground truth local coordinates by transforming the global ground truth data into each node's local coordinate system. Furthermore, at inference time, we also must transform the predicted local coordinates into global coordinates for the Kalman Filter to create one unifying prediction in the global coordinate system. The transformation between global distributions $\mathbf{\mu_g} \in \mathbb{R}^{3\times1}, \mathbf{\Sigma_g} \in \mathbb{R}^{3\times3}$ and local distributions $\mathbf{\mu_l}, \mathbf{\Sigma_l}$ is implemented using a translation (node position) $\mathbf{T}$ and a rotation $\mathbf{R}$ (node orientation):
\vspace{-5pt}
\begin{equation}
    \mathbf{\mu_g} = \mathbf{R} (\mathbf{\mu_l} - \mathbf{T}), \mathbf{\Sigma_g} = \mathbf{R} \mathbf{\Sigma_l} \mathbf{R}^T.
\end{equation}
\vspace{-10pt}



\noindent\textbf{Experiments and Results}
The model is trained on 83 minutes of data with 13 different viewpoints, calibrated with 5 minutes of data with three different viewpoints, and tested on 6 minutes of data with six different views. The model is trained and tested with \textcolor{black}{three configurations: RGB only, RGB + depth, and all-modalities. The training takes roughly 48-72 hours on a single RTX 4090 GPU, where including more modalities leads to a longer training time. }We aim to provide a baseline model for object tracking that is generalizable to different sensor placements without retraining. The results are shown in Table~\ref{tab:exp2}. We find that our model fails to produce a meaningful result using RGB images only, as the AD exceeds 120 cm within an area of $3 \mathrm{ m} \times 3.6 \mathrm{ m}$. 
However, when the depth modality is incorporated, we observe a significant improvement. The AD drops to approximately 60 cm, improving by 50\%. This demonstrates the potential of leveraging multimodality to overcome the limitations of relying solely on RGB images. The addition of mmWave and acoustic modalities further pushed the mean tracking error down to 47.48~\!cm. We can see that the inclusion of more sensing modalities generally makes the model more robust to the domain shift caused by sensor perspective changes. Despite the performance improvement brought by multimodal sensor fusion, an AD around 45 cm falls short of the desired accuracy. By releasing \name to the research community, we hope the dataset will contribute to the development of multimodal sensor fusion models more robust to sensor perspective shifts.



\vspace{-5pt}
\section{Limitations, Future Work, and Conclusions}
\label{Sec:final}

In this paper, we introduced \name, a multimodal, multiview dataset for indoor geospatial tracking using networked sensors. We showed baseline experiments where we could use \name to develop robust algorithms for adversarial sensing conditions such as poor illumination. We also demonstrated the potential to construct models agnostic to sensor placement variances. \textcolor{black}{Here, we discuss the limitations and possible future extensions of \name.}

\noindent\textbf{\textcolor{black}{Extension to other indoor layouts and scenarios.}} \textcolor{black}{Limited by the scale of the experiment site and the lack of IRB approval for collecting data in public space, \name is limited to a racetrack setting in a single room and does not cover complicated indoor layouts with obstacles and occlusions. We hope future studies will diversify \name by extending it to various realistic indoor scenarios.}\looseness=-1

\noindent\textbf{\textcolor{black}{Better multimodal sensor fusion architectures.} }While we show that the \name dataset enables the exploration of several research problems, our baseline results are still \textcolor{black}{preliminary}. In the single-view baseline, our early fusion model leans heavily towards only 1-2 modalities, limiting its robustness to missing sensor feeds. Our solution to the multiview baseline is also less satisfying in terms of absolute tracking error. Furthermore, our baselines do not cover the case of multiple tracking targets. These questions can be addressed in the future with more efforts towards better neural model designs.\looseness=-1

\noindent\textbf{\textcolor{black}{Incidents and accidents during data collections.}}
During the collection of \name, we have recorded many incidents such as car collisions, vehicles running off track, and humans rescuing stuck cars. \textcolor{black}{We included a field called $valid\_range$ in the metadata where we manually inspected the data and marked the data void before and after each session. If any incident causes the car to stay still for a long period, the corresponding part will be clipped out by applying the $valid\_range$ mask.} Meanwhile, we see these incidents as an opportunity to investigate other problems. For example, a learning-enabled agent may perform Q\&A about the experiment data, such as asking how many times the car has run out of lane or when a car collision occurred. Of course, doing so requires labeling incidents separately.\looseness=-1  

In conclusion, we hope \name opens up the opportunity to investigate widespread robustness issues in multimodal sensor fusion and to create lightweight, deployable models for indoor geospatial tracking, which can benefit the design and implementation of autonomous systems in smart building infrastructures.

\begin{acks}
The research reported in this paper was sponsored in part by: the IoBT REIGN Collaborative Research Alliance funded by the Army Research Laboratory (ARL) under Cooperative Agreement W911NF-17-2-0196;  the National Science Foundation (NSF) under award \#1822935; and the Air Force Office of Scientific Research (AFOSR) under award FA9550-22-1-0193. The views and conclusions contained in this document are those of the authors and should not be interpreted as representing the official policies, either expressed or implied, of the funding agencies.
\end{acks}

\bibliographystyle{ACM-Reference-Format}
\bibliography{citations}

\end{document}